**Distinguishing mirror from glass: A 'big data' approach to material perception.**

Hideki Tamura[1,2,*], Konrad E. Prokott[3], Roland W. Fleming[3]

[1]Department of Computer Science and Engineering, Toyohashi University of Technology, Toyohashi, Aichi, Japan
[2]Japan Society for Promotion of Sciences, Chiyoda, Tokyo, Japan
[3]Department of Experimental Psychology, Justus-Liebig-University Giessen, Giessen, Germany

*Corresponding author
Email: tamura13@vpac.cs.tut.ac.jp

*CLASSIFICATION*: **BIOLOGICAL SCIENCES (Neuroscience)**

Running title: **Distinguishing mirror from glass**

**Abstract**

Visually identifying materials is crucial for many tasks, yet material perception remains poorly understood. Distinguishing mirror from glass is particularly challenging as both materials derive their appearance from their surroundings, yet we rarely experience difficulties telling them apart. Here we took a 'big data' approach to uncovering the underlying visual cues and processes, leveraging recent advances in neural network models of vision. We trained thousands of convolutional neural networks on >750,000 simulated mirror and glass objects, and compared their performance with human judgments, as well as alternative classifiers based on 'hand-engineered' image features. For randomly chosen images, all classifiers and humans performed with high accuracy, and therefore correlated highly with one another. To tease the models apart, we then painstakingly assembled a diagnostic image set for which humans make highly systematic errors, allowing us to decouple accuracy from human-like performance. A large-scale, systematic search through feedforward neural architectures revealed that relatively shallow networks predicted human judgments better than any other models. However, surprisingly, no network correlated better than 0.6 with humans (below inter-human correlations). Thus, although the model sets new standards for simulating human vision in a challenging material perception task, the results cast doubt on recent claims that such architectures are generally good models of human vision.

**Significance Statement**

Recent breakthroughs using neural networks to solve challenging vision tasks like object recognition, have led to excitement about their potential as models of biological vision. Our findings urge caution. We show that neural networks can get good at distinguishing 'mirror' from 'glass' materials. However, to assess how similar models are to humans, it is not sufficient to compare accuracy. A good model should also predict the characteristic errors that humans make. We introduce a new method for doing so, by comparing performance on carefully selected diagnostic images with thousands of neural networks. The best models explain less than 30% of the variance in consistent human judgments, suggesting more care should be taken in using neural networks to model human vision.

Different materials, such as steel, silk, meat or glass, have distinctive visual appearances, and our ability to recognize such materials by sight is crucial for many tasks, from selecting food to effective tool use. Yet, material perception is challenging. The retinal image of a given object is the result of complex interactions between the object's optical properties, 3D shape and the incoming light (1–4). Thus, a given material can take on an enormous variety of different appearances depending on the lighting, object shape and viewpoint. At the same time, similar objects with different material properties can create quite similar images in terms of the raw spatial patterns of colour and intensity (5). To succeed at material perception, the visual system must somehow tease apart similar images belonging to different materials, while at the same time grouping together very diverse images belonging to the same material class (6, 7). This is a fundamental aspect of biological visual processing, which remains poorly understood.

A particularly challenging case is to distinguish polished mirror-like specular materials ('mirror') from colourless transparent materials ('glass') (8–13). Both kinds of material derive their appearance entirely from their surroundings, but through different light transport processes (**Figure S1**). Mirrors create a distorted reflection of the surrounding world, whereas for glass materials, incident light also enters the material, refracts and may reflect internally multiple times before re-emerging. Yet, in both cases, changing the object shape or surrounding world radically alters the image. As a result, the visual cues we use to distinguish between mirror and glass must generalize well across an enormous variety of images. At the same time, to distinguish the two kinds of material, the visual system must presumably use quite sophisticated image measurements that latch onto subtle differences in the image resulting from the way light interacts with them. Thus, investigating how the brain distinguishes mirror from glass can tap into the core processes underlying visual surface appearance more generally.

We reasoned that to work out how the visual system distinguishes mirror from glass, it is useful to take a 'big data' approach in which we embrace the enormous diversity of images of mirror and glass materials that confront the visual system. In particular, using computer graphics we sought to create a dataset of hundreds of thousands of images, which could then be data mined for reliable visual cues. To do the data mining, we turned to deep learning methods.

Over the last five or so years, artificial neural networks (14, 15) have demonstrated significant potential as models of biological vision (e.g., 16–21; see also reviews (22–27)). We set out to leverage these advances to gain insights into the visual processes underlying the challenging material perception task of distinguish mirror from glass. Comparisons between human vision and computational models typically use randomly selected images (20, 21, 28, 29), for which both humans and models achieve high performance. In contrast, our goal was to develop a model that would behave like humans according to more strictly defined criteria. Specifically, we sought a model that could not only predict the *successes* of human judgments, but also systematic *errors*, which are presumably the hallmarks of the processes unique to human visual computations. To do this, we created a 'diagnostic' image set that yielded systematic and consistent visual errors as well as correct percepts. However, to our surprise, we found that despite an extensive and guided search, none of the neural networks we investigated correlated better than 0.6 with human judgments.

**Results and Discussion**

*Predicting perception on randomly selected images*

Using computer graphics, we rendered over 750,000 images of mirror and glass objects (**Figure 1A** and **Stimuli**) of a wide variety of shapes, naturalistic illuminations and viewpoints. Half were mirror objects (pure specular reflection), the other half were glass (refractive index 1.5). Of these, we randomly selected 500 images of each material class and asked 10 volunteers to rate each image on a five-point scale, where 1 indicated that the object looked compellingly like glass, 5 indicated that it looked compellingly like a polished metal, and intervening values indicated different degrees of ambiguous appearance. **Figure 1B** shows a clear bimodal distribution of ratings, with mean ratings of 0.66 for mirror images and 0.19 for glass, and an accuracy of 77.9% correct responses. This suggests that observers are capable of distinguishing mirror from glass.

Before investigating deep learning models in depth, we tested the extent to which relatively simple image measurements could predict perceptual mirror/glass judgments (see **Figure S2A**). Specifically, we compared human performance with two 'hand-engineered' image-computable classifiers based on pixel intensity and color histograms ('Color-Hist') and 'mid-level' texture statistics (30) ('Port-Sim'). All classifier types were trained on half the dataset (chosen at random) and tested on the other half, and all images that were shown to humans were excluded from training and test sets. The Color-Hist classifier used eight pixel-histogram statistics (mean, variance, skewness, and kurtosis of luminance and saturation distributions). 'Port-Sim' used 1,052 features derived—via PCA—from both color and higher-order wavelet coefficient statistics (30); see **Methods**). These two classifiers were trained to classify mirror and glass with a logistic regression, with the output ranging from zero (glass) to one (mirror). Surprisingly, although based on quite simple image measurements, both of these classifiers achieved accuracies that almost rivaled human performance on the 1,000 images rated by our observers (**Figure 1C**). This suggests that despite the complex optics of reflection and refraction, there are many potential cues that would suffice to perform significantly above chance at distinguishing the two kinds of materials. However, we sought the specific cues that the human visual system relies on. A better test of this is the correlation between the Color-Hist and Port-Sim models and humans on an image-by-image basis. Although the models did correlate significantly with human performance, they did so significantly less well than individual humans do (Color-Hist vs humans: $t(18) = 6.056$, $p < 0.001$; Port-Sim vs humans: $t(18) = 2.356$, $p < 0.05$, t-test), suggesting that humans do not rely on the same cues as these simple classifiers (**Figure 1D**).

As an initial attempt to investigate the potential of CNNs to predict human performance, we trained 10 instantiations of a feedforward network with three convolution layers (see **Figure S2B** and **Methods**). The 10 instances had identical architecture, but different initial random weights and randomly selected training images (2-fold cross validation). Note that the random images for human psychophysics (**Figure 1B**) were not used in this training. On the same random images as before, the networks achieved an accuracy that superseded humans, and correlated with mean human responses within the same range as individual humans did, thus outperforming the two 'hand-engineered' classifiers. This suggests that CNNs learn features that are inherently superior to the simple color and texture statistics. This in itself is unsurprising as the CNNs learn many more features (99,410), and thus perform the classification in a higher-dimensional space. However, for the purposes of understanding biological vision, the key question is whether the features learnt by the CNNs resemble those used by the human visual system.

To gain further insights into the nature of the internal representations of the classifiers, we then performed representational similarity analysis (RSA) (31) using the images that had been rated by humans. **Figure 2A** shows the Representational Dissimilarity Matrices (RDMs; (31)) for each of the classifiers, as well as ground truth. The rows/columns of the matrix represent the different images, ordered into two blocks by their true class (mirror vs. glass) and within a block by their mean human ratings (from most mirror-like to most glass-like). Individual entries represent the dissimilarity between the corresponding pair of images in terms of the perceived or predicted mirror vs. glass ratings (see **Methods**). Thus, low values indicate the corresponding pair of images are represented as highly similar, while higher values indicate they are more dissimilar.

The patterns in the matrices suggest that for these randomly selected images, humans and classifiers broadly agree. We can summarize the relationships between the RDMs in a Classifier

Correlation Matrix (CCM; also known as a 'second-order RDM' (31)), in which each row/column indicates a different observer or computational classifier, and each entry contains the mean correlation between the RDMs for the corresponding pair of observers/models (**Figure 2B**). For comparison, we also included 10 random RDMs, to characterize how much more similar the classifiers are to humans than would occur by chance. Applying multidimensional scaling (MDS) to the CCM allows us to visualize the relationships in 3D (**Figure 2C**). The correlation between each classifier and humans were 0.30 (Color-Hist), 0.31 (Port-Sim), and 0.58 (CNN), respectively. This reveals that all three classifier types learn inter-image relations that are significantly and substantially closer to humans than occurs by chance, and of all classifier types, the CNNs appear to acquire the most similar representation to humans. These results tend to suggest that feedforward convolution neural networks have significant potential as models of human visual judgments of mirror and glass materials.

*Creating a dataset of images diagnostic of human vision*

Based on the high correlations between observers and the computational models, it could be tempting to conclude that the models accurately simulate human visual processes. However, there are several reasons for caution. First, the main purpose of comparing models based on different features is to identify which features best predict human material perception. Yet, for randomly selected images, even the most primitive models appear to match human perception quite well. Given what we know about early vision and material perception (32–34), it seems highly unlikely that visual perception of mirror *vs* glass is based on raw luminance and color distributions, which are entirely insensitive to the spatial structure of the image. Second, and more importantly, it is possible that the high correlations simply result from the fact that both humans and classifiers achieve quite high accuracies. If all models correctly assign most images to one of the two distinct modes ('mirror' or 'glass'), then it follows that they will tend to correlate with one another. Indeed, in **Figure 2A**, 58% of the variance in the human judgments is accounted for by the ground truth. A good model must be able to predict not only the successes of human vision, but also the specific pattern of errors, on an image-by-image basis. To test this, we need a set of diagnostic images that decouples accuracy from human judgments.

Creating such a dataset is nontrivial as most images are perceived correctly. It is not sufficient to identify images for which participants are inconsistent in their interpretation, as a deterministic, image-computable model cannot even in principle account for variations between observers when presented with the same image. Our goal is to predict that proportion of the variance in judgments which is *consistent* across observers, and therefore we need an image set that includes images that are consistently misperceived. Specifically, the goal was to identify a 'benchmark' set of images with a flat distribution across the five bins ranging from 'mirror' to 'glass' ratings, for both mirror and glass images (in contrast to the skewed distributions for random images in **Figure 1B**), thereby perfectly decorrelating the true material class from the perceived class. We set as our criterion of consistency that ten out of ten naïve observers should rate each image in the same bin of the 5-point rating scale.

To obtain this diagnostic image set, we performed a sequence of experiments using both crowd sourcing and laboratory judgments to progressively funnel images down to a set that is highly diagnostic of human performance (**Figure 3A**; see **Figure S3** and **Methods** for details). Identifying images that are perceived veridically is relatively straightforward, so we prioritized identifying 'illusory' images that are misperceived, reasoning that the bins corresponding to veridical percepts could be filled in afterwards. Specifically, from the full set of renderings, we selected 30,000 images at random, 1,500 of which were presented to each of 20 observers. Based on the responses, we then selected about 11,000 images—sampling uniformly from the

judgments—which proceeded to the next round, in which images were rated by more observers via crowdsourcing. From their responses, we selected the 522 images which had been consistently rated as belonging to the wrong class by at least three observers. In the final round, ten observers rated each of these images, resulting in a total of 102 images which had been consistently rated as belonging to non-veridical bins. To fill in the veridical bins of the distribution we then had another ten observers rate 1,000 randomly chosen images from the renderings. From the responses, we selected 68 images that were consistently perceived correctly by all ten observers, yielding 170 images, i.e., 34 images in each of the five bins from perceived mirror to perceived glass, half of which were actually mirror and half glass. Some examples are shown in **Figure 3B** (see **Figure S4** for details).

To increase the number of diagnostic images, we also trained a generative adversarial network (GAN; (35, 36)) on our renderings. GANs consist of a generator network, G, that is trained to produce images, which a discriminator network, D, has to distinguish from a given dataset. During training D improves at distinguishing the synthesized images from the training data, while G learns to create images that are hard to discriminate from the training data. The result is a model that can synthesize images with many of the visual characteristics in common with the renderings (see **Figure 3B**, also **Figure S4B** for details). We find that such images include many cases that are more ambiguous than renderings, appearing somewhere between mirror and glass. In another 'funnel' sequence of experiments, we identified 95 images (out a set of 1,400), which were consistently rated by ten observers as belonging to specific bins. Combining the selected GAN images with the renderings yielded a total 265 diagnostic images, in which true material class was perfectly decorrelated from perception (**Figure S4A**).

*Systematic exploration of the space of feedforward networks*

With this diagnostic image set in hand, we then sought neural network models that would correlate strongly with humans for these diagnostic images. It is important to note that the space of potential convolution network models is very large: they can vary widely in terms of their architectures, hyperparameters and training schedules. We reasoned that within the space of feedforward neural networks, some networks are likely to approximate human visual processing better than others. We therefore ran a large-scale search through the space of feedforward networks with the general form depicted in **Figure 3C**, varying the network depth systematically (see also **Figure S5A**). All networks consisted of an input layer followed by a basic 'block' of layers comprised of convolution, batch normalization (37), ReLU (38), and max pooling layers, which were repeated several times, followed by dropout (39), fully connected and softmax layers, and ending with a two-unit classification output ('mirror' vs 'glass'). To investigate the effect of architecture depth, we systematically varied the number of basic blocks in the sequence from one to twelve (See Methods). Then, for each network depth, we ran 200 iterations of Bayesian hyperparameter search (BHS) to identify the values of 11 hyper-parameters controlling the network architecture and training (e.g., number of filters per layer, initial learning rate, momentum; see **Figure S5B**) in an 'optimization-stage'. The objective of the BHS was to optimize correlation to human judgments on the diagnostic image set, which progressively improved across iterations (**Figure 3D**). All networks were trained on a randomly chosen 400,000 renderings (200,000 images in each class) with 2-fold cross validation in order to converge the network quickly.

Having identified promising hyperparameters for each architecture depth, in the 'validation-stage', we then trained 30 instances of each of the resulting neural networks (differing only in the initial random state), again using the same number of renderings, with half for training and the other half for testing. Importantly, these networks were never trained on the diagnostic images, and training proceeded until the validation accuracy had not improved for at least three validations, independently for each architecture depth. The mean correlations between the networks and humans on the diagnostic image set is shown in **Figure 4**. Of all depths, the 3-layer network architecture ('OptCNN') was the one that correlated best with human judgments, and was the network class we considered for further analysis. Importantly, however, none of the thousands of networks trained throughout the BHS or the final validation exceeded a correlation with humans of 0.54 whereas human-to-human correlations was between 0.61 and 0.81. In other words, although OptCNN was the closest of all models we considered, it still failed to capture average human performance as well as even the most unrepresentative of the individual humans did.

**Figures 4A** and **4B** compare the highest of the OptCNN networks to humans and the other classifiers on the diagnostic image set. By design, humans perform at chance on these images (**Figure 4A**). All classifiers outperform humans in terms of accuracy, yet OptCNN is the closest, making the most errors on these images, even though it was trained with the same objective function and a very similar training set to the original CNN, which performs too well to resemble humans. In terms of image-by-image correlations to human judgments (**Figure 4B**), none of the classifiers reaches the noise ceiling, but OptCNN significantly outperformed all other classifiers (Color-Hist vs OptCNN: $t(9) = 4.79 \times 10^{13}$, $p < 0.001$; Port-Sim vs OptCNN: $t(9) = 5.19$, $p < 0.001$; CNN vs OptCNN: $t(9) = 20.23$, $p < 0.001$, t-test).

To compare the nature of the representations in the different classifiers and humans, we then performed RSA. To do this, we measured the similarity between all images in the diagnostic image set according to the final classification output of each classifier. To visualize the relationships between the different classifiers, we computed the mean correlation between the different classifier types, and then performed MDS (as in **Figure 2D**). **Figure 4D** shows the different classifier types arranged in the first three resulting dimensions. This analysis reveals that OptCNN was the most similar to humans, although, there is still a substantial residual difference.

The greater similarity between OptCNN and the default CNN is also revealed by a more detailed view into the representations at different processing stages of the networks. We applied RSA using the diagnostic image set, to the input stage, the three ReLU stages after each convolution layer, the fully connected layer (FC) and the final output in both the original CNN and OptCNN. We then computed correlations between each of the resulting RDMs ('Layer Correlation Matrix', LCM), and performed MDS to visualize the relationships between the different representations, along with the true labels and human judgments (**Figure 5A**). At the input stages of the network, images that are perceived by humans as mirror and glass materials are thoroughly entangled, such that further processing is required to separate them. As we proceed through network layers, we see that the representations in CNN and OptCNN diverge. While CNN's representations increasingly approach the ground truth, OptCNN's representation increasingly organizes images in the way that humans do, such that in the output stages, images that subjectively appear like mirror or glass are teased apart. It is interesting to note that the representation in the original CNN's fully connected layer to approach human performance. At this stage of the network, glass and mirror are still substantially entangled, suggesting, again that human judgments are not optimal given the data available in the images.

Robustness to noise is another key characteristic of human vision (40). We find that OptCNN outperforms the original CNN in terms of the effects of noise on the correlation with human performance. If we perturb the input images with noise, the networks' predictions about the material tend to change (**Figure 5B**). Importantly, we find that while the correlation between CNN's predictions and the human judgments of the unperturbed images falls precipitously as noise is added, for OptCNN, not only is the correlation higher across all noise levels, the decline is also gentler. This suggests that by identifying networks that more closely resemble humans in terms of their solution to the objective function, we also identify representations that capture other aspects of human perception, such as robustness to noise.

**General Discussion**

As many materials that we can easily recognize did not exist until the last few centuries—or even decades—our ability to recognize them must be learned rather than evolved. How the visual system acquires the visual computations and internal representations that allow us to succeed at material perception remains poorly understood. Here, we investigated the extent to which deep learning can reveal representations that resemble human judgments.

Studies comparing humans to machine learning models often focus on overall performance at a task (20, 28, 29, 41–43), or correlation on arbitrary images (20, 29, 41–43) for which both humans and successful machine vision system tend to perform well. Here, by contrast, we defined a new criterion for comparing neural networks with human judgments, by creating a *diagnostic* set of images in which human performance is decorrelated from ground truth. This allows us search for networks that capture the characteristic eccentricities of human vision, reproducing the tell-tale errors that humans tend to make. Although identifying such an image set is time consuming and effortful, it provides a benchmark against which all future models of human vision can be tested. Here, we created one such set for a challenging material perception task: the discrimination of mirror and glass materials.

By comparing human performance on the diagnostic image set against an array of models, we were able to show that neither simple colour statistics, nor more sophisticated texture statistics can predict human judgments of mirror vs glass. More importantly, we also showed that an arbitrarily-defined CNN, which appeared to resemble humans when tested on arbitrary images, did not resemble humans very closely at all when evaluated on the diagnostic image set. The use of arbitrary design decisions is widespread in the literature comparing neural networks to brain activity or human behaviour. Our findings suggest that more care should be taking in explicitly searching for, or fitting neural network models to biological data.

To do this, we then performed a large-scale, systematic search through the space of feedforward networks trained to distinguish mirror from glass objects, in search of a neural network that more closely resembled human performance. The network architecture that performed most similarly to humans (OptCNN) was a three-layer network, which for arbitrary images was not especially good at distinguishing mirror from glass in terms of the true physical labels (at least compared to rival models). This suggests that humans are far from optimal at the task; indeed, many artificial neural networks can outperform them. However, importantly, and to our surprise, even the optimised model did not reproduce average human behaviour on the diagnostic image set as well as individual humans do. While previous studies have noted that neural networks have reached the 'noise ceiling' of human-to-human performance (e.g., 20, 42), we find that when tested on a test set that is truly diagnostic of human vision, such conclusions might not in fact be warranted, and that further research is required to find good models of human vision.

Why did OptCNN fail to match human performance? There are at least three important respects in which the models differ from humans and the human brain. First, one of the most striking aspects of human visual cortical processing is the massive amount of feedback (44–46). It is widely believed that feedforward processing is responsible for many of our visual abilities. For example, high-level visual tasks, such as animal detection (47) can be successfully completed too rapidly for feedback to contribute substantially. Nevertheless, given its anatomical extent, feedback presumably plays an important role in visual processing, potentially in selective visual attention, visual imagery and the learning process that establishes the representations in the first place. Here, we considered only feedforward architectures. It could be that a key missing ingredient in OptCNN is recurrent processing, and that adding feedback signal flow could make up some of the shortfall in correlation with humans. Feedback might, for example, be necessary for performing long-range spatial computations, such as comparing the structures within the region of the object with those of the background.

A second important difference is the nature of the training objective. Here—as in almost all neural network-based putative models of human vision (23, 26, 27)—we used *supervised learning* in which the network is trained on hundreds of thousands of accurately labelled images. Human vision is unlikely to be trained this way, as feedback is very sparse, and the scale of the training set almost certainly exceeds human visual experience with mirror and glass objects. In particular, we very rarely get to see mirror and glass versions of the same objects, and we presumably also exploit the fact that vision unfolds continuously over time, rather than in independent static snapshots, as CNNs are typically trained (although see 48, 49). It is much more likely that visual representations are learned through *unsupervised processes*, and this may have a critical effect on the internal representations that the visual system learns.

A third important difference between the artificial neural networks and humans lies in the nature of the task that the networks are trained on. Human vision is not tailored solely to the task of distinguishing mirror from glass objects, whereas here, we trained the networks on a binary classification, effectively separating the entire world into two possible states. The representations that optimize performance on this task may well be quite general purpose, as has been found with neural networks optimized for object recognition, which can easily be repurposed for other tasks, such as action recognition (50) and image semantic segmentation (51). Nevertheless, it is also possible that in being trained on such a constrained task, the networks learned representations that do not resemble human visual processes.

Future work should use a combination of unsupervised learning, more naturalistic objective functions and network architectures that more closely resemble primate cortex to tease these possibilities apart. We have shown that for most images, even arbitrarily designed artificial neural networks outperform more conventional 'hand-engineered' models, and thus have substantial potential as models of human visual processes. Nevertheless, when their similarity to humans is investigated with a stricter criterion of predicting the specific patterns of errors humans make, they still have important shortcomings. Although neural networks can be found or fitted to the brain or human behaviour, they should not yet be seen so much as 'out of the box' accurate working model of human brain processes, but rather as an experimental platform for further research, much as animal models of neurological disorders are.

**Methods**

*Observers*

*Lab experiments (i.e., not 'crowdsourcing')*: 60 observers were students of Justus-Liebig-University Giessen (JLU) and Toyohashi University of Technology (TUT) with normal or corrected-to-normal vision. All experimental protocols were approved by the Ethics board at Justus-Liebig-University Giessen and were conducted in accordance with the Code of Ethics of the World Medical Association (Declaration of Helsinki). Observers at JLU were paid 8 Euro / hour, those at TUT were not paid. Informed consent was obtained from all observers.

*Crowdsourcing*: 247 participants were recruited via the Clickworker platform and were paid 1.2 Euro each. Before the beginning of the experiment participants were presented with an online consent form that explained the purpose and procedure of the experiment, as well as the uses and benefits of their participation. All participants that took part in the experiment agreed to these conditions and that their data be recorded and stored anonymously for research and publication in scientific journals.

*Stimuli*

*Renderings:* Images were rendered using Mitsuba renderer (52). We selected 1,583 objects from Evermotion (https://evermotion.org) and 253 high dynamic range light fields for the illumination from the Southampton-York Natural Scenes (SYNS) dataset (53) and the other sources (54, 55) (see also supplement information). For the 'mirror' objects, the BSDF was a 'conductor' model with 100% specular reflectance. For the 'glass' material, BSDF was 'dielectric', with internal refractive index of 1.5. Objects were uniformly scaled to fit within the unit sphere, and placed at the origin. The camera was randomly located at a position between 30 and 60 degrees elevation angle and any azimuth with a constant distance of 2 units in Mitsuba. The sampling count was 512 per pixel with the Sobol Quasi-Monte Carlo sampler. The reconstruction filter was set as Gaussian. The renderer generated the final image, at $256 \times 256$ pixel resolution with gamma correction (56). Then, they were resized to $64 \times 64$. Note that mirror and glass images were paired using same object and illumination but different camera locations to avoid that the classifiers simply learn a pixel difference between mirror and glass images. We screened all images and excluded a small number of images with rendering artifacts. The final dataset contains 753,696 images.

*GAN images:* Two generative adversarial networks, GANs (35, 36) were trained to synthesize images that they could not distinguish from a given training set of renderings. Specifically, the rendered images (see Renderings) were split into two subsets (mirror and glass) as the training sets (376,848 images in each). The network architecture and the hyper parameters were the same as a previous network (36), except for the minor modifications in the standard Tensorflow DCGAN implementation that avoid the discriminator of the network converging too fast (57). After 20 epochs training, we then generated 700 images from each GAN by inputting random noise vectors to create a total of 1,400 images, which were then rated by humans.

*Apparatus*

Stimuli were displayed on a 27-inch liquid crystal display (Eizo CG276 at TUT and CG277 at JLU) using factory default settings with a resolution of 1920 × 1200 pixels. Stimulus presentation was controlled by MATLAB using Psychtoolbox 3.0 (58–60).

*Experiment 1: Random renderings*

Ten observers participated (7 women; age range: 19 to 38 years; average 25.1 ± 5.2 years) in the lab. We randomly selected 1,000 images from the dataset (500 mirror and 500 glass images) and presented them in random order to each observer (i.e., each image was rated by all 10 observers). The images subtended NNN degrees visual angle and were separated by a distance of NNN cm. They were presented on a uniform gray background with a fixation cross in the center of the screen. The observers were asked to rate each stimulus on a five-point scaling (glass to mirror) by pressing a corresponding key on the keyboard. They could respond at any time, but the stimulus disappeared after one second.

*Experiment 2: Diagnostic images*

The purpose of this experiment was to create a 'benchmark' set of diagnostic images, with

(1) a uniform distribution of appearances ranging from mirror to glass; (2) perceptual appearance that is decorrelated from the true material class ('ground truth') and (3) consistent judgments across observers. Identifying images that correlate with ground truth (upper left and lower right quadrants of matrix in **Figure 3B**) is straightforward, as humans are generally good at distinguishing mirror vs glass for our renderings. Thus, most of the procedure revolves around finding images that systematically yield errors (i.e., the upper right and lower left quadrants of **Figure 3B**). To achieve this, we ran two parallel series of experiments using renderings (series A) and images generated by GANs (series B), respectively. Each series starts with a large number of images, with images being progressively excluded in each round, to arrive at a much smaller final set covering the desired distribution (see also **Figure S3 and S4**).

*Round A1 (Rendering ratings)*: Twenty observers (all men; age range: 21 to 26 years; average 23.1 ± 1.4 years) participated in the lab. We randomly selected 30,000 renderings from the dataset (50% mirror, 50% glass) and distributed 1,500 images to each observer. The procedure was the same as in experiment 1, except that the task was a three-way judgment ('mirror', 'glass', or 'hard to recognize'). The last option was used to exclude rendering artifacts for further rounds (2.9% of the images were excluded here). **Figure S6A** shows results of this round. In total, 10,976 images moved ahead to round A2. Specifically, 2,744 images were randomly selected from each of the four bins other than the 'hard to recognize' images (i.e., mirror that looks like either mirror or glass and glass that looks like either mirror or glass).

*Round A2 (Rendering ratings)*: 247 crowdsourced participants observed the stimuli selected by round A1, and were asked to rate them on a five-point scale (glass to mirror). They were each shown 100 images—98 randomly chosen test images from the output of round A1 and 2 catch trial images, consisting of photographs with clear mirror or glass appearance. Only the 5586 images that were rated by at least three crowd-workers were analyzed further. **Figure S6B** shows rating results of this round. Based on the responses, we selected 522 images, in which the ratings conflicted with the ground truth material, by selecting from the two 'outermost' bins of the distribution for each class. Specifically, 261 mirror images with rating score 0.0–0.4 (i.e., seen as glass) and 261 glass images with rating score 0.6–1.0 (i.e,. seen as mirror). These images progressed to round A3.

*Round A3 (Rendering ratings)*: Ten observers participated in the lab (9 women; age range: 21 to 30 years; average 24.8 $\pm$ 2.8 years). The procedure was the same as in experiment 1 (5-bin rating task). The experiment consisted of 1,566 trials (3 trials × 522 images from round A2), and all trials were randomly ordered. **Figure S6C** shows results of this round. From these, a total 102 images were selected for the diagnostic image set, by selecting from the three 'outermost' bins from each class. Specifically, 51 mirror images with ratings 0.0–0.6 (i.e., seen as glass or ambiguous) and 51 glass images with ratings 0.4–1.0 (i.e., seen as mirror or ambiguous). These selected images were included in the diagnostic image set.

*Round B1 (GAN-image screening)*: Some GAN-generated images resemble textures rather than objects with distinct material properties. The purpose of Round B1 was to exclude such images from subsequent rounds. Ten observers participated in the lab (8 women; age range: 20 to 32 years (average 24.8 ± 4.1 years). The stimuli were 1,400 images generated by GANs (see *GAN images*). The procedure was the same as in experiment 1, except that the task was to indicate in a binary decision whether the object shape and material were recognizable or not). **Figure S7A** shows results of this round. Based on the responses, 500 images that were judged to be recognizable by at least six out of ten observers were moved ahead to Round B2.

*Round B2 (GAN-image ratings)*: Ten observers participated in the lab (all women; age range: from 21 to 34 years; average 25.1 ± 3.8 years). The stimuli were 560 images including 500 GAN images from Round B1 and 60 renderings (30 mirror and 30 glass images) from round A2, which had received ratings that were highly consistent with ground truth. The procedure was the same as in experiment 1 (rating task). The experiment was composed of 1,680 trials (3 trial × 560 images from Rounds B1 and A2), and all trials were randomly ordered. **Figure S7B** shows result of this round. We selected 95 images (19 images from each bin) to add to the diagnostic image set.

*Final Diagnostic Image Set*: The two streams of experiments resulted in a final diagnostic image set of 265 images including both mirror and glass renderings, along with GAN images with prediction score uniformly distributed from 0.0 to 1.0 (**Figure 3B** and **Figure S3**). These are composed of 68 veridical images (from experiment 1), 102 illusory images (from Round A1-A3), and 95 GANs' images (from Round B1 and B2).

*'Hand-Engineered' Classifiers*

We developed three different classifiers (**Figure S2A**): Color-Hist, Port-Sim, and a CNN with manually selected hyperparameters. Color-Hist used eight features: mean, variance, skewness and kurtosis of intensity and color saturation from 64 × 64 RGB image. The features of Color-Hist were z-scored across all images. To get the features of Port-Sim, we first used the texture analysis algorithm of Portilla and Simoncelli (30) to extract 3,381 higher-order image statistics. These were z-scored and the number of dimensions reduced to 1,052 by principal component analysis (cumulative explained variance of complete image set > 99%). For both Color-Hist and Port-Sim, a logistic regression was trained to distinguish mirror from glass based on the ground truth labels. CNN was defined as a three-layer convolutional neural network with 64 × 64 RGB image input and the binary (mirror *vs*. glass) classification output. The network architecture and training hyperparameters are shown in **Figure S2B**. All classifiers were trained and tested with 2-fold cross validation, which was repeated 10 times with different randomly selected training and test sets (images that were shown to human were excluded here). The final output—a prediction score ranging from zero (as glass) to one (as mirror)—was averaged across training repetitions.

*Identifying optimal CNN models through Bayesian hyperparameter search*

We used Bayesian hyperparameter search through a space of feedforward architectures to identify which CNNs correlated best with humans (using MATLAB R2017b with Neural Network Toolbox and Statistics and Machine Learning Toolbox; **Figure S5A**). The objective was to maximize the correlation coefficient between CNN and human on the diagnostic image set. The network architectures were basically the same as the CNN in experiment 1, except that we parametrically varied the 'depth', i.e., the number of layers from 1 to 12, by repeating a block of layers consisting of convolution, batch normalization, rectified linear unit, and max pooling layers before the first fully connected layer. Note that the max pooling layers were only used up to 3 layers (the last 3 layers) because of the size constraints of the filters. For each depth, we ran 200 iterations of the Bayesian hyperparameter search (i.e., 200 CNNs were generated with different hyperparameters, in search of the optimal values for each depth; **Figure S5B**). Each CNN was trained and tested with same training and test set. Having identified the optimal hyperparameter values, we then trained 30 new CNNs with those optimal parameters, but with different random initial weights and training/test images. These are the networks that are reported in **Figure 3C**.

*Representational similarity analysis (RSA)*

We defined two different representational dissimilarity matrices (RDM), a 1st-stage RDM to identify dissimilarity relationships between images in humans and each classifier; and a 2nd-stage RDM ('Classifier Correlation Matrix', CCM), characterizing how similar the 1st-stage RDMs are across different humans and classifiers, allowing us to compare their internal representations. We also defined a 'Layer Correlation Matrix' (LCM) to compare different layers within the CNN and OptCNN networks. The 1st-stage RDM was defined as Euclidean distance of prediction scores (final output) from each classifier or average of observers' response from human. The 2nd-stage RDM was defined as a dissimilarity, which was one minus Pearson's correlation between each 1st-stage RDMs.

The number of dimensions of the 2nd-stage RDM was reduced to three or two dimensions using MDS and we visualized a relationship between the targets as three-dimensional space in **Figure 2C** (1,000 images from the random image set) and **Figure 4D** (265 images from the diagnostic image set), and as two-dimensional space in **Figure 5A** (265 images from the diagnostic image set).

**Acknowledgments**

This work was supported by JSPS KAKENHI Grant Number JP16J00273, the DFG (SFB-TRR-135: "Cardinal Mechanisms of Perception", project number 222641018), and an ERC Consolidator Award (ERC-2015-CoG-682859: "SHAPE"). The authors wish to thank Shigeki Nakauchi for enabling the exchange visits necessary for this project and Karl Gegenfurtner for comments on an earlier version of the manuscript.

25. LeCun Y, Bengio Y, Hinton G (2015) Deep learning. *Nature* 521(7553):436–444.
26. Majaj NJ, Pelli DG (2018) Deep learning—Using machine learning to study biological vision. *J Vis* 18(13):1–13.
27. Yamins DLK, DiCarlo JJ (2016) Using goal-driven deep learning models to understand sensory cortex. *Nat Neurosci* 19(3):356–365.
28. Ghodrati M, Farzmahdi A, Rajaei K, Ebrahimpour R, Khaligh-Razavi S-M (2014) Feedforward object-vision models only tolerate small image variations compared to human. *Front Comput Neurosci* 8:74.
29. Kheradpisheh SR, Ghodrati M, Ganjtabesh M, Masquelier T (2016) Humans and Deep Networks Largely Agree on Which Kinds of Variation Make Object Recognition Harder. *Front Comput Neurosci* 10:92.
30. Portilla J, Simoncelli EP (2000) A Parametric Texture Model Based on Joint Statistics of Complex Wavelet Coefficients. *Int J Comput Vis* 40(1):49–71.
31. Kriegeskorte N, Mur M, Bandettini P (2008) Representational similarity analysis - connecting the branches of systems neuroscience. *Front Syst Neurosci* 2:4.
32. Anderson BL, Kim J (2009) Image statistics do not explain the perception of gloss and lightness. *J Vis* 9(11):1–17.
33. Kim J, Anderson BL (2010) Image statistics and the perception of surface gloss and lightness. *J Vis* 10(9):1–17.
34. Marlow PJ, Kim J, Anderson BL (2012) The perception and misperception of specular surface reflectance. *Curr Biol* 22(20):1909–1913.
35. Goodfellow I, et al. (2014) Generative Adversarial Nets. *Advances in Neural Information Processing Systems 27*, pp 2672–2680.
36. Radford A, Metz L, Chintala S (2015) DCGAN: Unsupervised Representation Learning with Deep Convolutional Generative Adversarial Networks. *arXiv* 1511.06434.
37. Ioffe S, Szegedy C (2015) Batch Normalization: Accelerating Deep Network Training by Reducing Internal Covariate Shift. *arXiv* 1502.03167.
38. Glorot X, Bordes A, Bengio Y (2011) Deep Sparse Rectifier Neural Networks. *Proceedings of the Fourteenth International Conference on Artificial Intelligence and Statistics*, pp 315–323.
39. Srivastava N, Hinton G, Krizhevsky A, Sutskever I, Salakhutdinov R (2014) Dropout: A Simple Way to Prevent Neural Networks from Overfitting. *J Mach Learn Res* 15:1929–1958.
40. Geirhos R, et al. (2017) Comparing deep neural networks against humans: object recognition when the signal gets weaker. *arXiv* 1706.06969.
41. Hong H, Yamins DLK, Majaj NJ, DiCarlo JJ (2016) Explicit information for category-orthogonal object properties increases along the ventral stream. *Nat Neurosci* 19(4):613–622.
42. Kubilius J, Bracci S, Op de Beeck HP (2016) Deep Neural Networks as a Computational Model for Human Shape Sensitivity. *PLOS Comput Biol* 12(4):e1004896.
43. Majaj NJ, Hong H, Solomon EA, DiCarlo JJ (2015) Simple Learned Weighted Sums of Inferior Temporal Neuronal Firing Rates Accurately Predict Human Core Object Recognition Performance. *J Neurosci* 35(39):13402–13418.
44. Budd JML (1998) Extrastriate feedback to primary visual cortex in primates: a quantitative analysis of connectivity. *Proc R Soc London Ser B Biol Sci* 265(1400):1037–1044.
45. Felleman DJ, Van Essen DC (1991) Distributed hierarchical processing in the primate cerebral cortex. *Cereb Cortex* 1(1):1–47.
46. Muckli L, Petro LS (2013) Network interactions: non-geniculate input to V1. *Curr Opin Neurobiol* 23(2):195–201.
47. Thorpe S, Fize D, Marlot C (1996) Speed of processing in the human visual system. *Nature* 381(6582):520–522.
48. Karpathy A, et al. (2014) Large-Scale Video Classification with Convolutional Neural Networks. *2014 IEEE Conference on Computer Vision and Pattern Recognition*, pp 1725–1732.
49. van Assen JJ, Nishida S, Fleming RW (2018) Estimating perceived viscosity of liquids with neural networks. *41st European Conference on Visual Perception (ECVP2018)*.
26

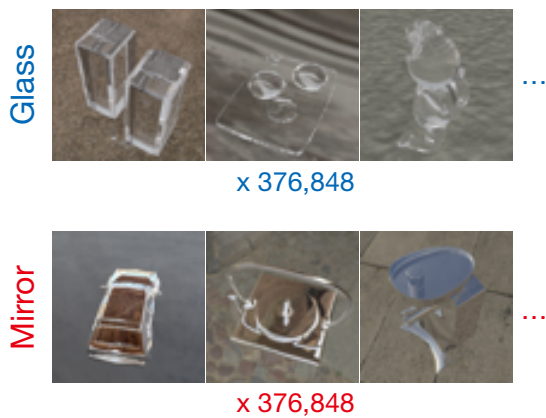
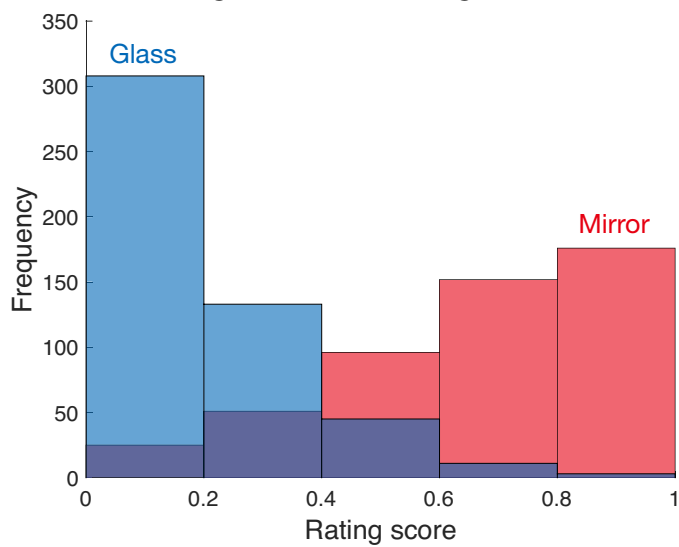
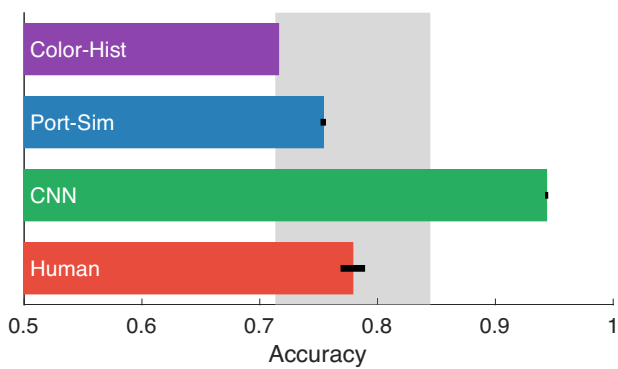
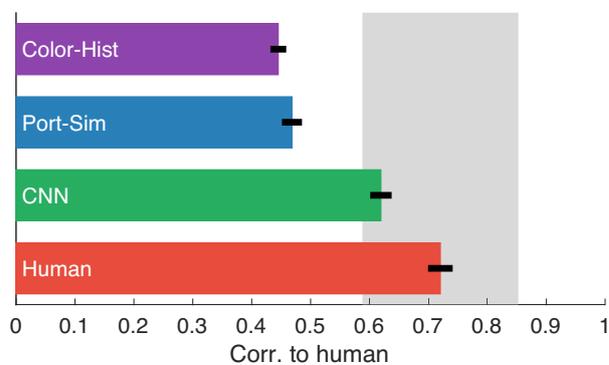

Figure 1. Results of randomly selected renderings

(A) Randomly selected example renderings from the data set. *Top row*: glass; *bottom row*: mirror. (B) Results of rating experiment with randomly selected images. Horizontal axis indicates rating score from glass to mirror (0-1). Vertical axis indicates a frequency of ratings in each bin across 10 observers. (C) Accuracy of human and model classifier responses for the test stimuli in experiment (mean of 10 repetitions in each classifier or 10 observers). Error bars represent standard error of the mean (error bar of Color-Hist is too small to see at this scale). Gray area indicates mean±2SD of all human observers. (D) Correlation coefficient between human and model classifiers for the test stimuli. Human-to-human correlation was defined as the average of 10 correlations between each observer and the mean of the remaining observers.

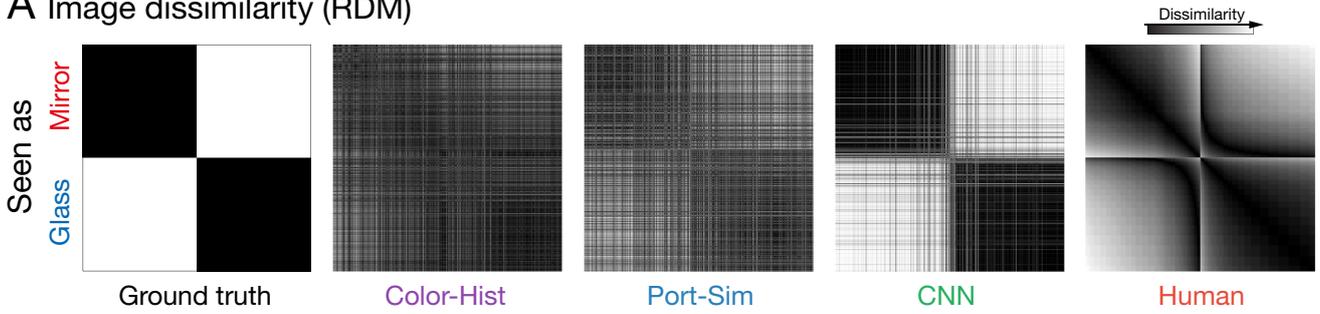
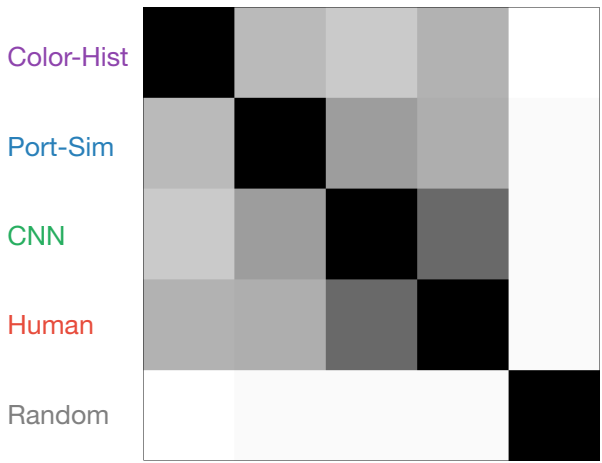
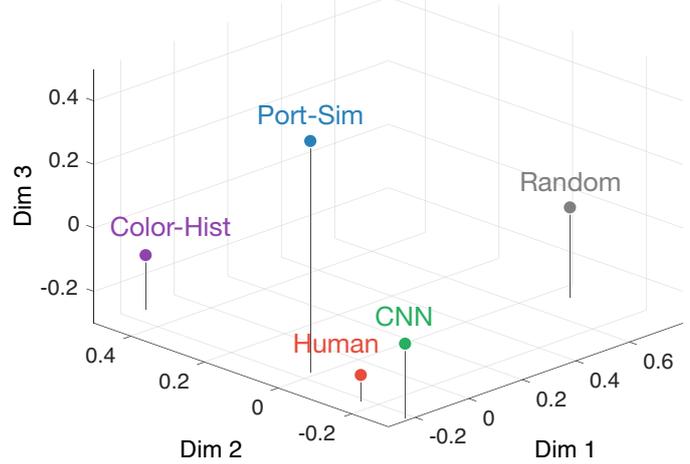

Figure 2. RSA of randomly selected renderings

(A) RDMs for ground truth, each classifier and human judgments. The rows/columns of each matrix represent 1,000 images, ordered into two blocks by their true class (mirror vs. glass) and within a block by their mean human ratings from most mirror-like to most glass-like. Individual entries represent the dissimilarity between the corresponding pair of images in terms of the perceived or predicted mirror vs. glass ratings. Darker entries indicate that the corresponding images are estimated to be highly similar, whereas brighter entries indicate they are more dissimilar. (B) CDM between models / humans. Each row/column indicates a different classifier, human, and random RDM as a control. Each entry contains the mean correlation between the RDMs for the corresponding pair of observers/models (intensity code as in A). (C) 3D visualization of the relationship between models/humans by applying MDS to the CDM. The three axes indicate the first three dimensions obtained by MDS.

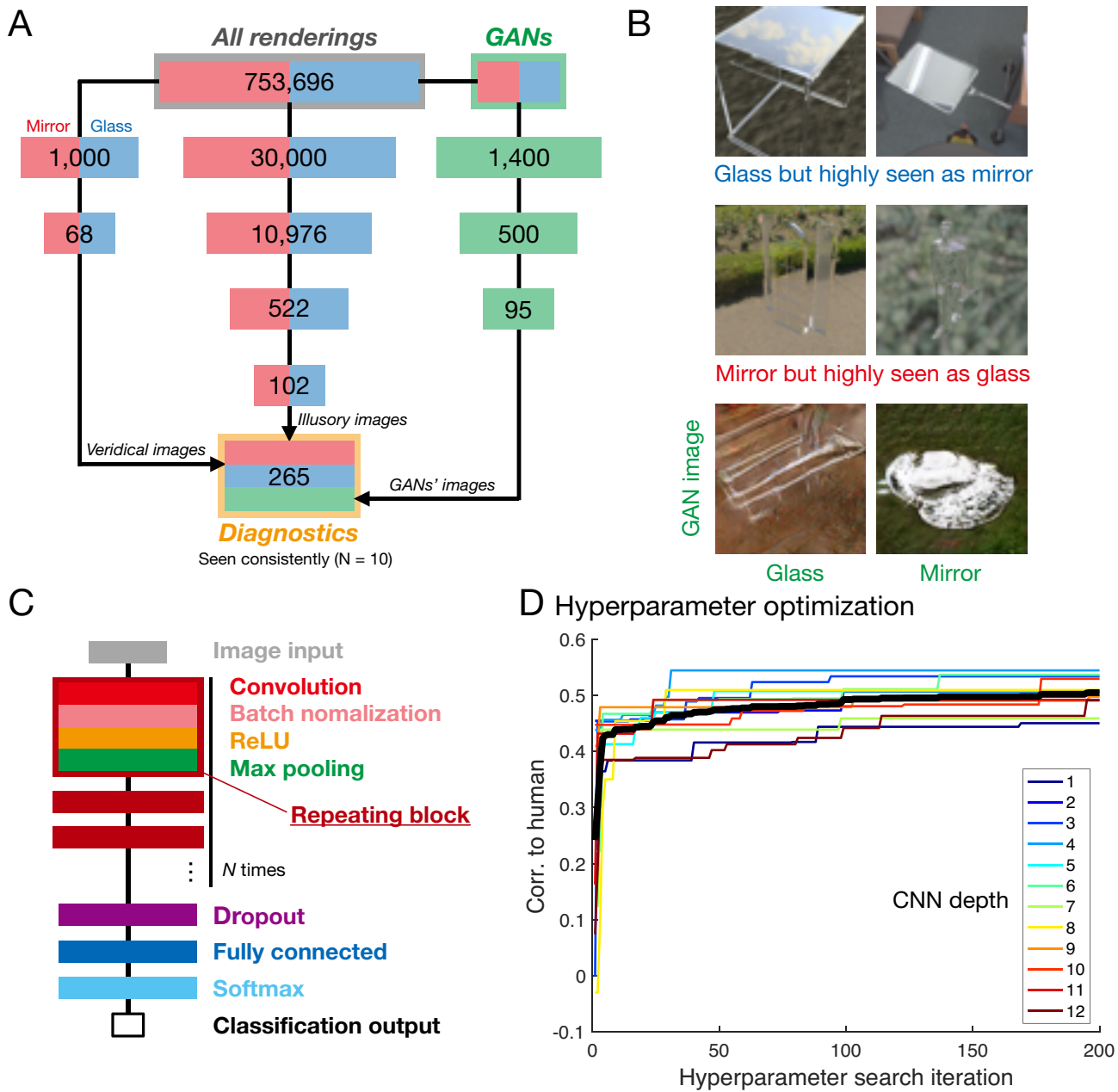

Figure 3. Diagnostic image set and systematic exploration of the space of the feedforward networks

(A) Summary flowchart showing creation of the diagnostic image set, which consists of three different image types: 'veridical' renderings (human judgments match ground truth), 'illusory' renderings (humans consistently misjudge material class), and 'GAN' images. See **Methods** and **Figure S5** for details of selection process. (B) Examples of illusory and GAN images from the diagnostic image set. *Top row*: glass objects that are seen as mirror; *middle row*: mirror renderings that are seen as glass; *bottom row*: GAN images that look like glass (left) and mirror (right). (C) General form of the feedforward network architecture in this study (see also **Figure S2**). (D) Results of Bayesian Hyperparamter Search (BHS). Horizontal axis indicates the number of iterations of BHS. Vertical axis indicates correlation between human to each model with different network depth (indicated by color, from 1 to 12). Thick black line represents mean of all 12 depths.

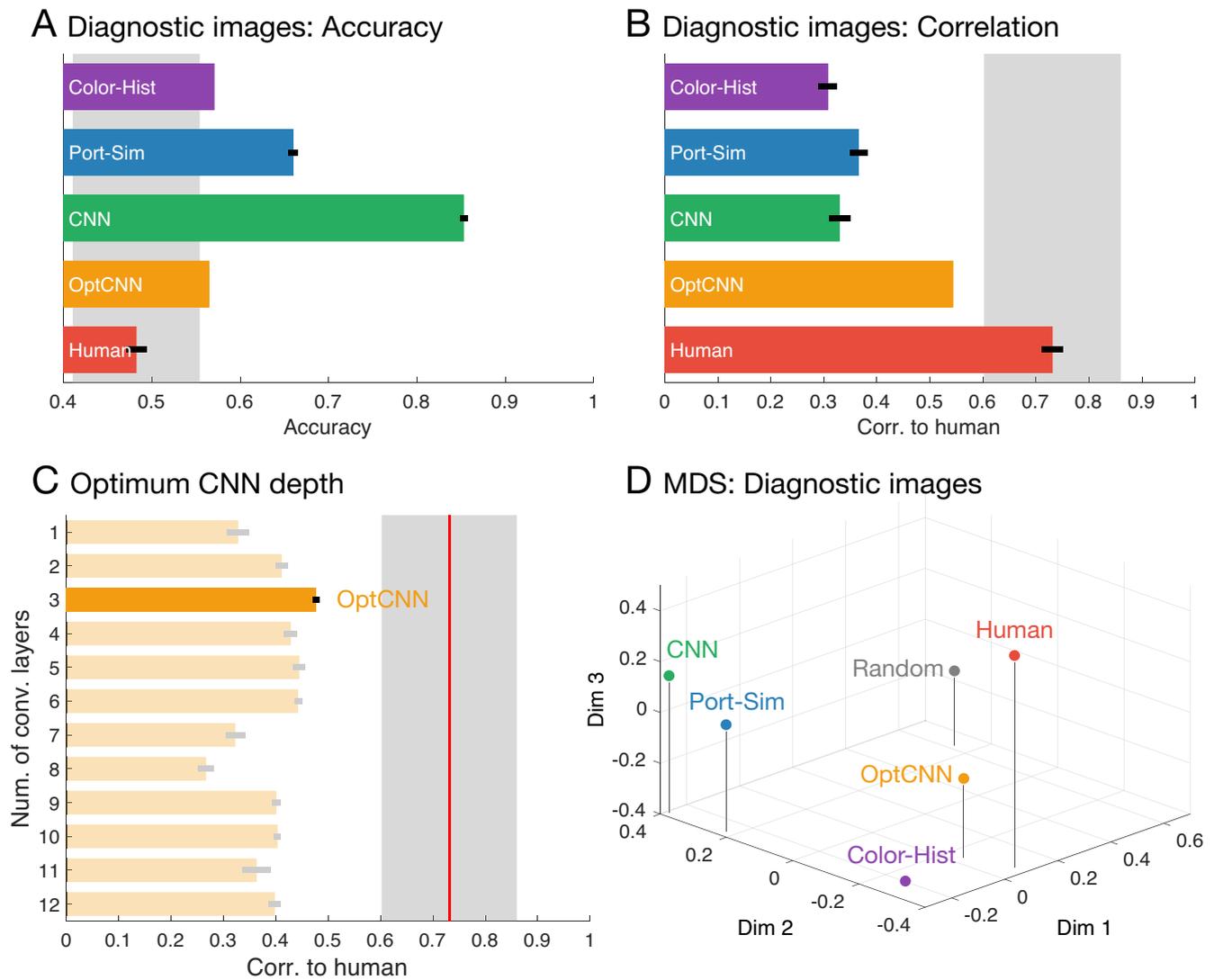

Figure 4. Results of the diagnostic image set

(A) Accuracy of humans and model classifiers for diagnostic image set. (B) Correlation coefficient between human and classifiers for diagnostic stimuli (Symbols same as **Figure 1C** and **1D**). OptCNN represents the highest correlation of 30 instances of 3-layer CNN from Bayesian hyperparameter search. (C) Correlation to humans for each network depth. Horizontal axis indicates correlation coefficient. Vertical axis indicates the number of convolution layers (i.e. the number of repeating blocks) in the networks. Red line and gray area indicate mean±2SD of all human observers. (D) 3D visualization of the relationship between the models via MDS (as in **Figure 2C**, but based on diagnostic image set).

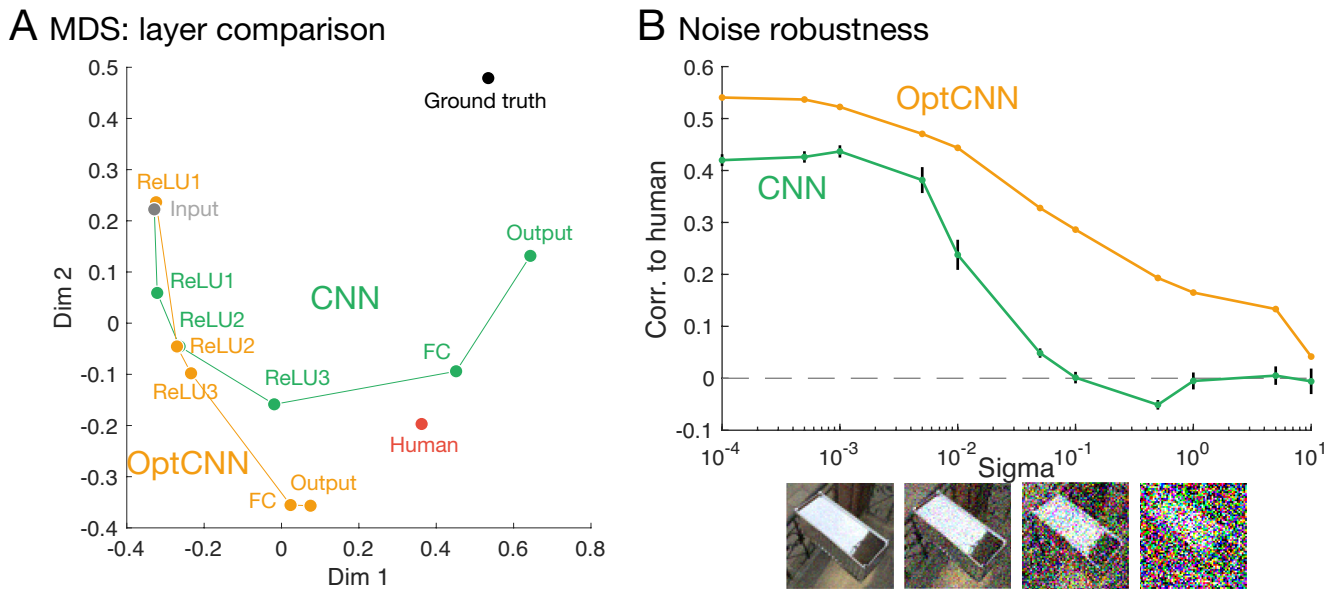

Figure 5. Comparing CNN and OptCNN in terms of the internal representation and robustness to noise

(A) 2D visualization of network internal representations between the layers (derived via MDS applied to the layer correlation matrices). The horizontal and vertical axes indicate the first two dimensions obtained by MDS. Two main streams show the relationship between layers in CNN (one of 10 instances, chosen randomly) and OptCNN (of 30 instances, the best correlating with human) with human judgements and the ground truth (see main text). (B) Robustness to noise. Horizontal axis indicates sigma of Gaussian noise (i.e., the amount of image perturbation). Four images show examples with different sigma ($10^{-3}$, $10^{-2}$, $10^{-1}$, and $10^{0}$). Vertical axis indicates the correlation to humans. Error bars represent standard error of the mean across all 10 classifiers (CNN). Note that OptCNN represents the highest correlation of 30 instances (same as OptCNN in A).

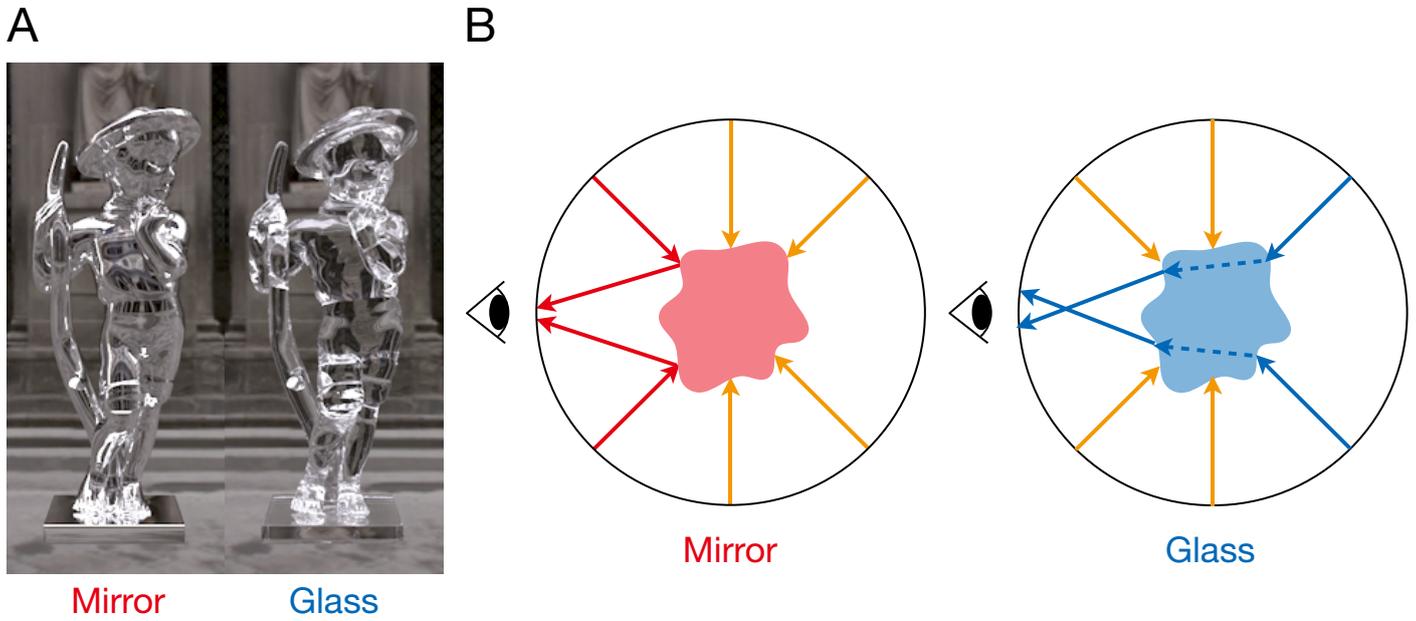

Figure S1. Distinguishing mirror from glass

(A) Example objects made of mirror (left) and glass (right) materials. 3D shape, illumination, and camera position are identical but the object's optical properties are different. (B) Illustration of different light paths through mirror and glass objects. Mirror reflects from the surface; glass refracts through the body of the object.

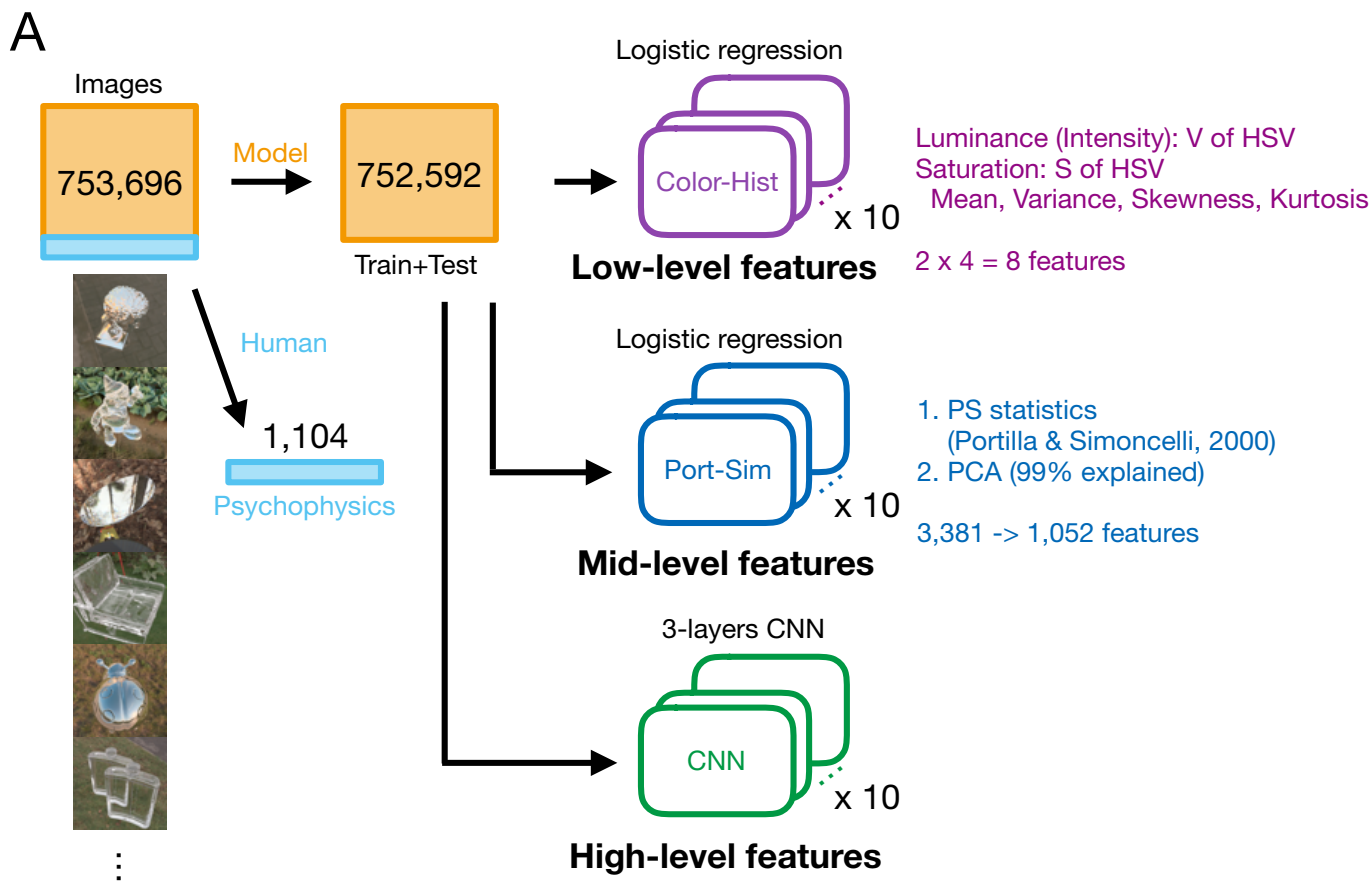

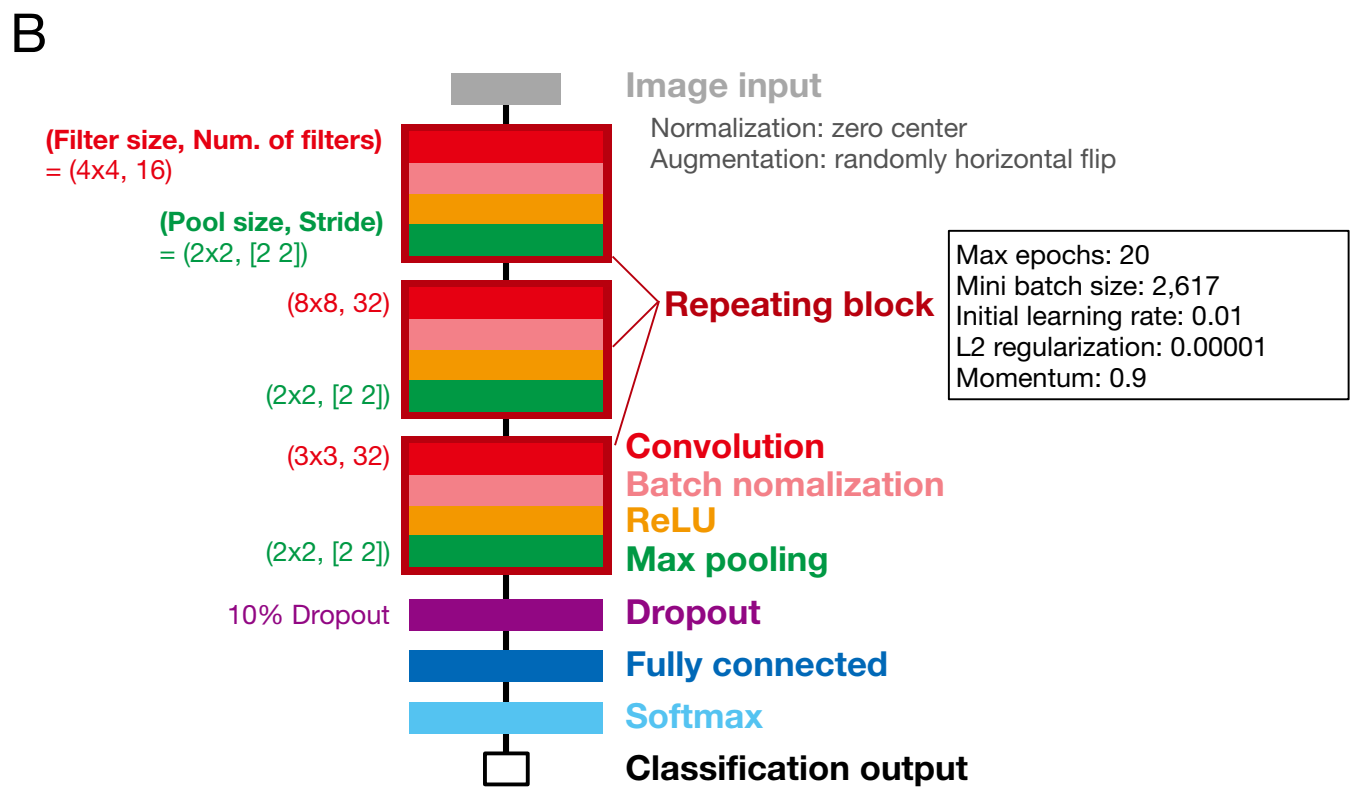

Figure S2. Classifiers and CNN architecture

(A) Flowchart showing development of three classifiers (Color-Hist; Port-Sim and CNN). (B) Network architecture of CNN. The text box shows the hyperparameters for training the CNN. The other hyperparameters were the same as default settings in MATLAB.

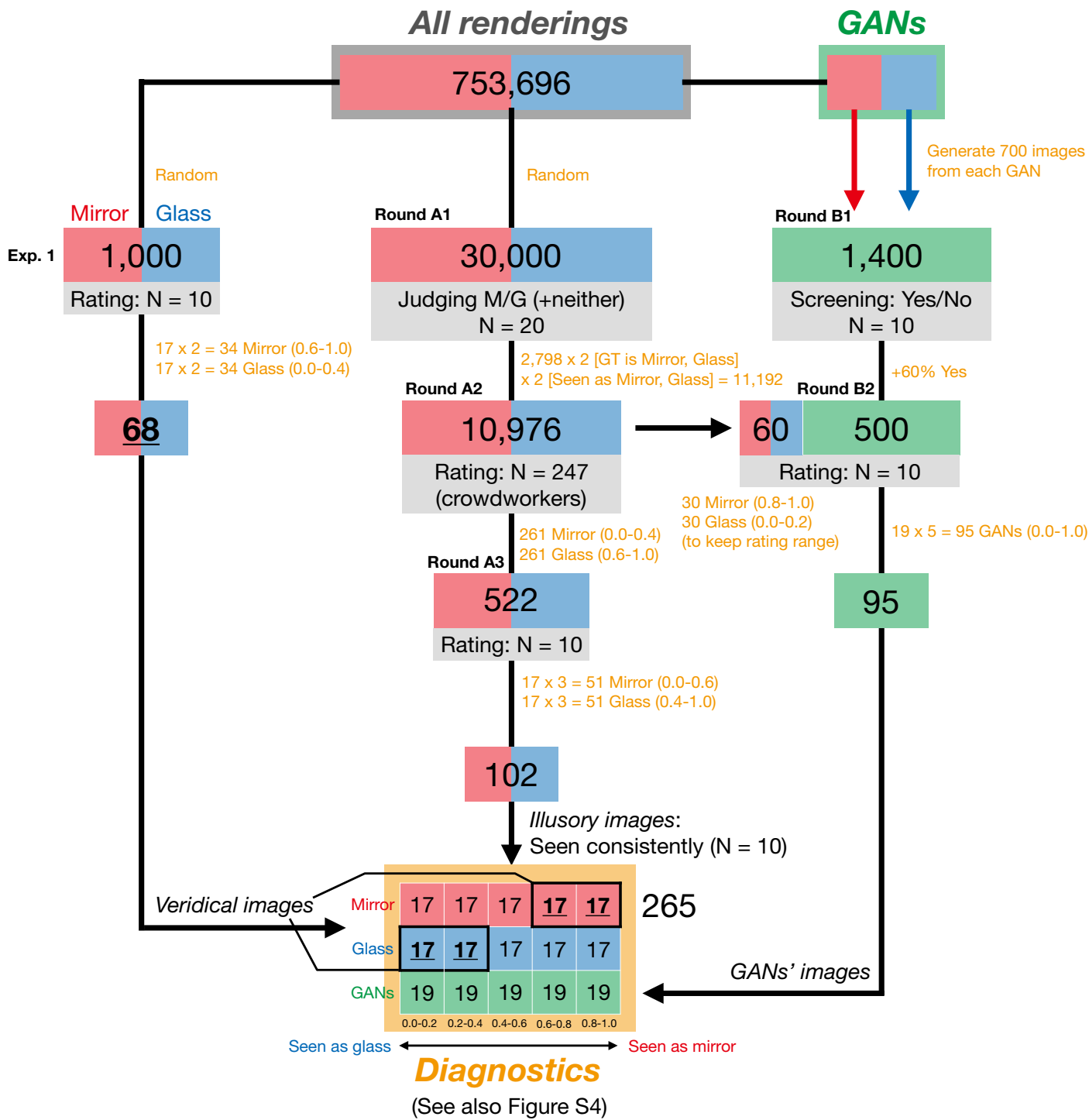

Figure S3. Detailed flowchart describing creation of diagnostic image set. Red indicates renderings depicting mirror materials; blue indicates renderings of glass materials; green indicates images generated using GANs. Each stage represents a different experiment used to select images for the subsequent stage in the corresponding sequence (A or B). See also **Figure S4** and Methods.

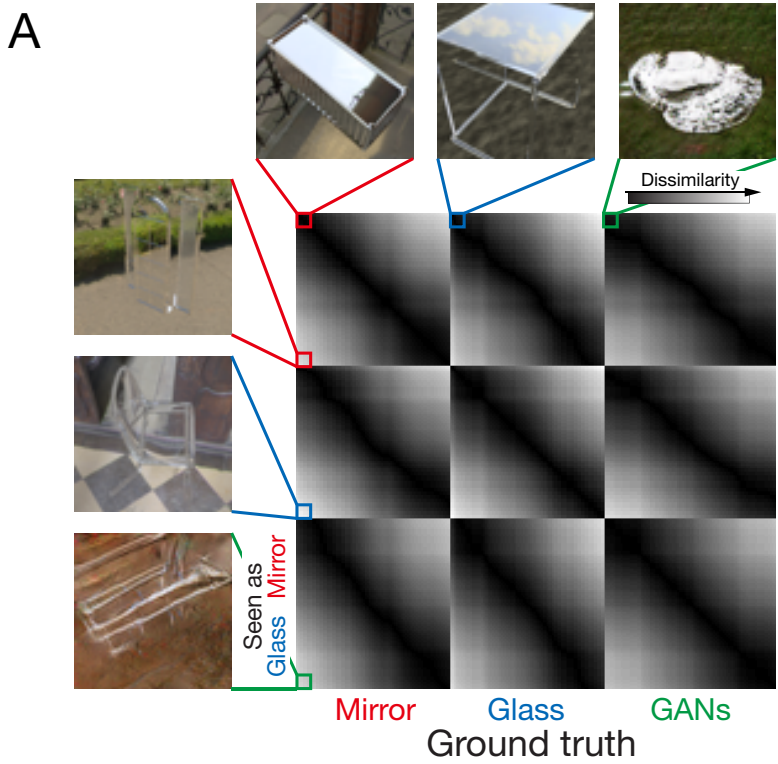

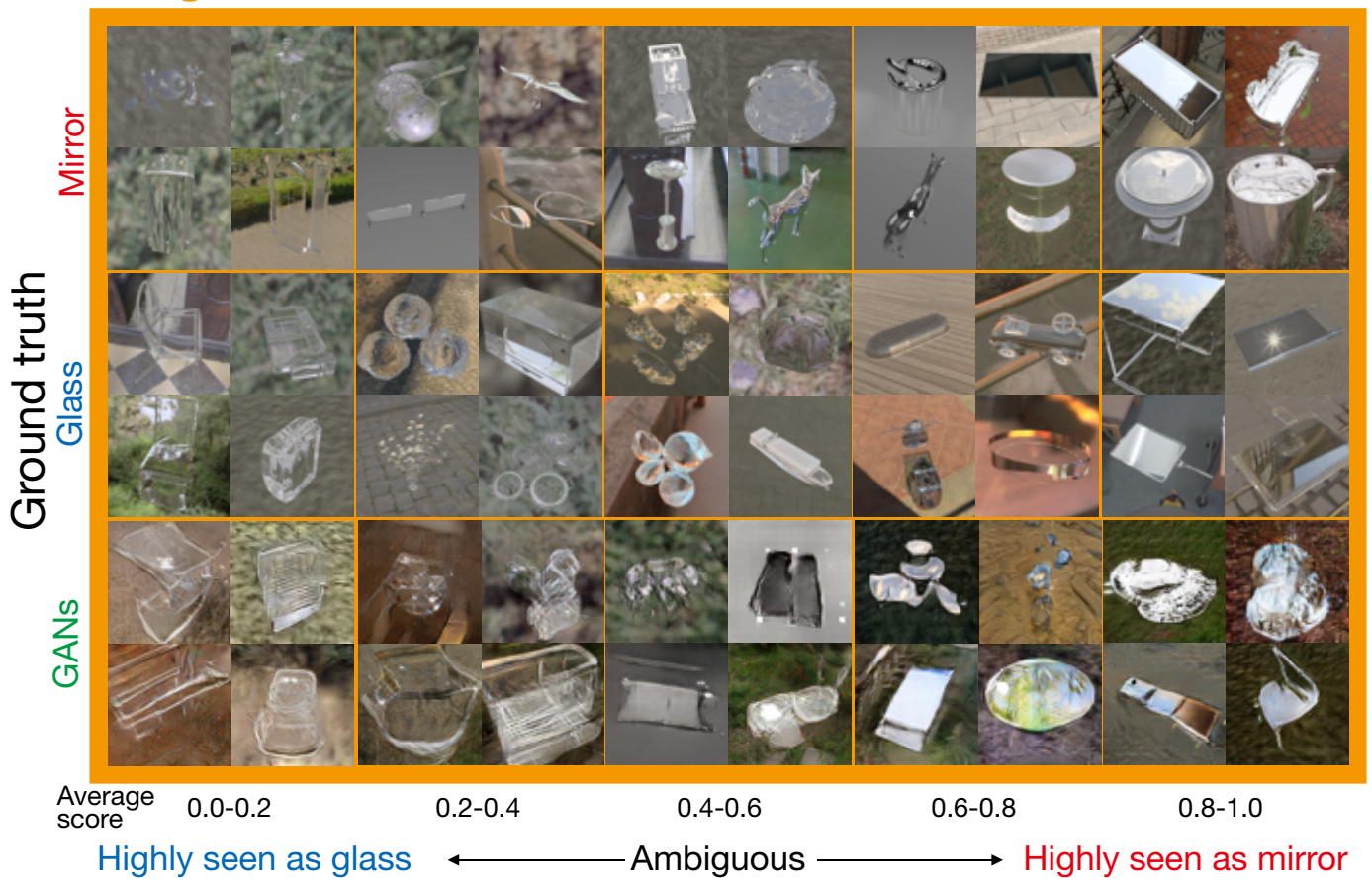

Figure S4. Example images from the diagnostic image set

(A) RDM of diagnostic image set. The format is the same as **Figure 2A** except adding GANs as the third class. The panel shows six example images with extremely high or low rating score in each class. (B) Example images. Each row indicates different ground truth (mirror, glass, and GANs). Each column indicates average rating score of 10 observers. See also **Figure S3**.

A

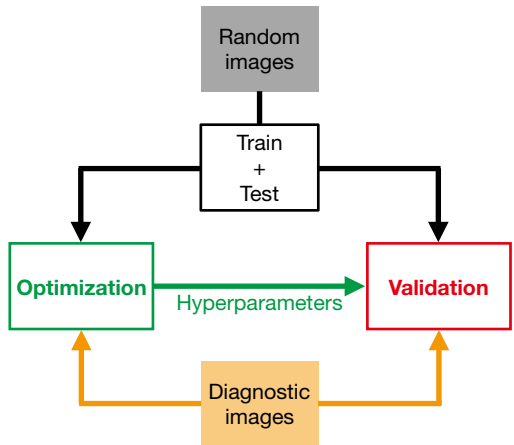

B Optimized hyperparameters

|  | Name | Min | Max | Type | Transform |
|---|---|---|---|---|---|
| 1 | Initial learning rate | $10^{-7}$ | $10^{-1}$ | Real | Log |
| 2 | Factor for $L_2$ regularization | $10^{-7}$ | $10^{-1}$ | Real | Log |
| 3 | Mini batch size | 100 | 2000 | Real | - |
| 4 | Momentum | 0.9 | 0.96 | Real | - |
| 5 | Dropout rate | 0 | 0.3 | Real | - |
| 6 | Filter size 1 | 2 | 3 | Integer | - |
| 7 | Filter size 2 | 2 | 4 | Integer | - |
| 8 | Filter size 3 | 2 | 4 | Integer | - |
| 9 | Num. of filters 1 | 4 | 8 | Integer | - |
| 10 | Num. of filters 2 | 4 | 30 | Integer | - |
| 11 | Num. of filters 3 | 4 | 30 | Integer | - |

Figure S5. Systematic exploration of the space of feedforward networks

(A) Illustration of optimization and validation stages of network exploration. Random renderings are used for training and test during training at both stages. Diagnostic images provide the objective of the Bayesian hyperparameter search, and for testing the trained networks at validation stage. (B) Eleven hyperparameters controlling the network architecture that were adjusted during optimization stage. Each hyperparameter was searched in the range between 'min' and 'max'. The first two hyper-parameters were transformed to log space during the searching. A pair of filter size 1 and num. of filters 1 was set to the convolution layers from the 1st to the n-2 th repeating block. A pair of filter size 2 and num. of filters 2, and a pair of filter size 3 and num. of filters 3 were set to the convolution layer in the penultimate and last repeating blocks. The learning rate was decreased by 0.1 times in each 10 epochs.

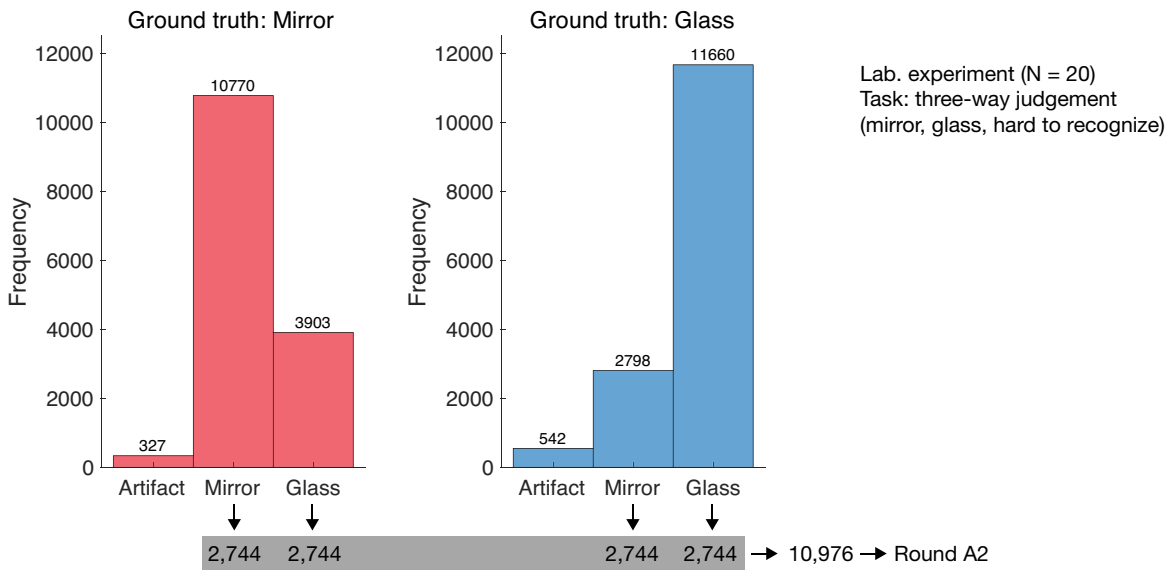
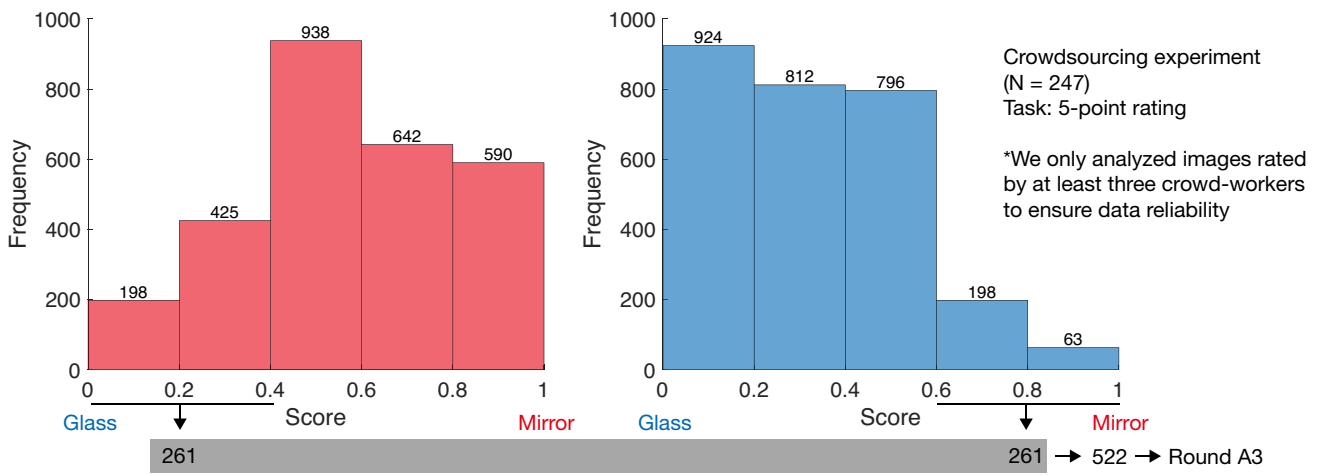
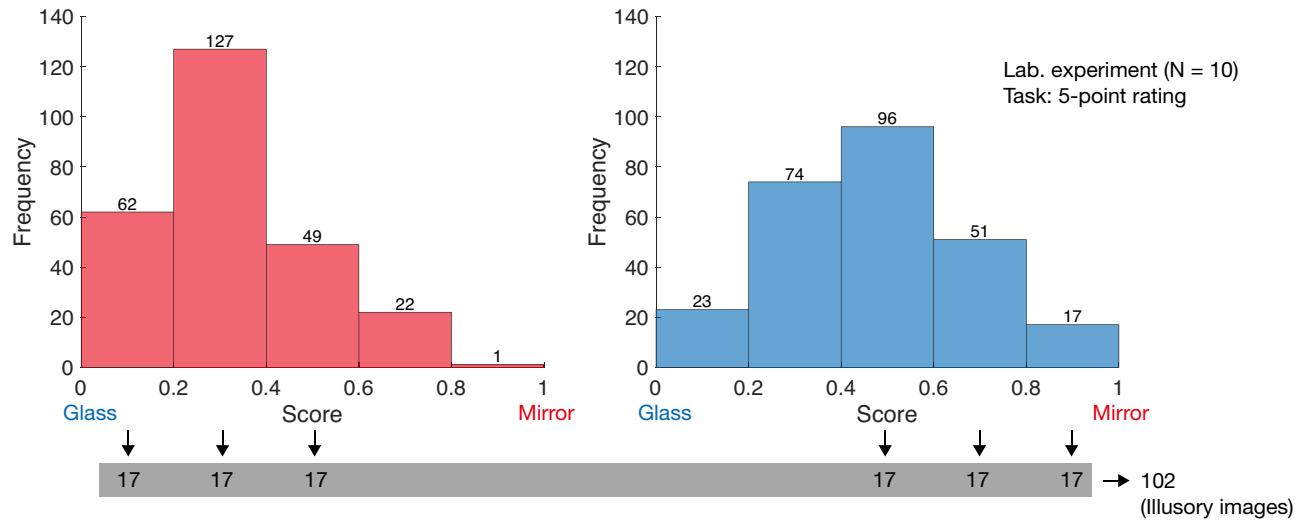

Figure S6. Results of Round A1-A3

Each panel shows the results of each round of experiments as histogram, as in **Figure 1B**. Red bars indicate ground truth mirror renderings; blue indicate ground truth glass renderings. Gray strips indicate number of items from each bin selected for subsequent round. See also **Methods**.

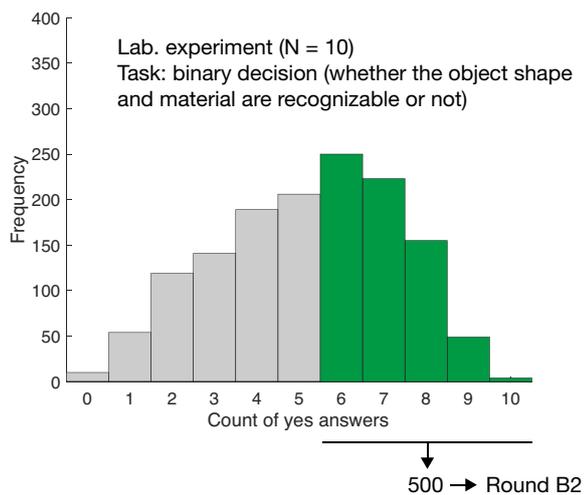 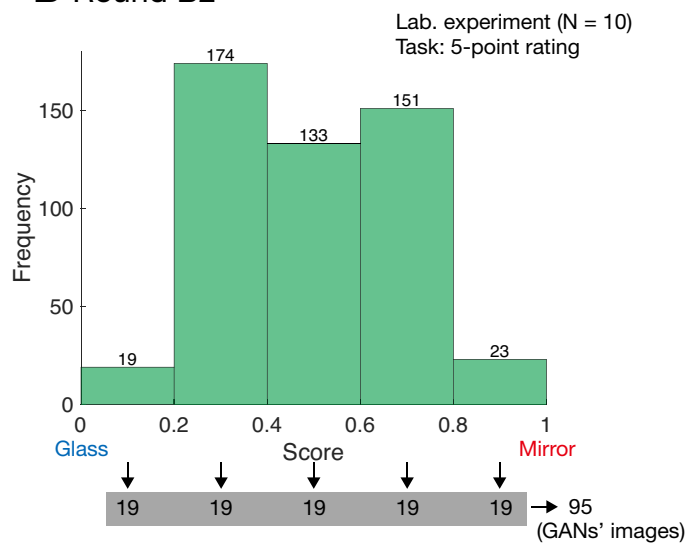

Figure S7. Results of Round B1-B2

Each panel shows result of each round as histogram of images. Green bars indicate GAN images (no ground truth label, unlike in Figure S6). Gray strips indicate number of images progressing to subsequent round. See also **Methods**.

# Supplement information

**Illuminations**

https://syns.soton.ac.uk/

http://www.pauldebevec.com/

http://hdrmaps.com/freebies

http://dativ.at/lightprobes/

http://www.openfootage.net/?cat=15

https://hdrihaven.com/hdris.php?thumb=all&sort=date&search=all&page=2&npp=12

https://www.doschdesign.com/